\documentclass[sigplan,10pt,screen,nonacm]{acmart}

\renewcommand\footnotetextcopyrightpermission[1]{}
\usepackage{booktabs}       
\usepackage{graphicx}       
\usepackage{subcaption}     
\usepackage{algorithm}      
\usepackage{algpseudocode}  
\usepackage{xspace}         
\usepackage{balance}        
\usepackage[table]{xcolor}  

\definecolor{rankfirst}{RGB}{31,120,180}   
\definecolor{ranksecond}{RGB}{166,206,227} 
\definecolor{rankthird}{RGB}{224,224,224}  

\newcommand{\best}[1]{\textbf{#1}}

\graphicspath{{figures/}{images/}}

\newcommand{\sysname}{the proposed framework\xspace}

\begin{document}

\title{Native LLM and MLLM Inference at Scale on Apple Silicon}

\author{Wayner Barrios}
\affiliation{%
  \institution{Wiqonn Technologies}
  \country{Colombia}
}
\email{wbarriosq@wiqonn.com}

\begin{abstract}
The growing adoption of Apple Silicon for machine learning development has created demand for efficient inference solutions that leverage its unique unified memory architecture. However, existing tools either lack native optimization (PyTorch MPS) or focus solely on text models, leaving multimodal workloads underserved. We present vllm-mlx, a framework for efficient LLM and MLLM inference on Apple Silicon built natively on MLX. For text models, we achieve 21\% to 87\% higher throughput than llama.cpp across models ranging from Qwen3-0.6B to Nemotron-30B, while providing continuous batching that scales to 4.3x aggregate throughput at 16 concurrent requests. For multimodal models, we introduce content-based prefix caching that eliminates redundant vision encoding by identifying identical images through content hashing, regardless of input format. Our evaluation on Apple M4 Max demonstrates throughput of up to 525 tokens per second on text models and 28x speedup on repeated image queries, reducing multimodal latency from 21.7 seconds to under 1 second. Video analysis with up to 64 frames achieves 24.7x cache speedup. We release our implementation as open source to support efficient inference on consumer Apple hardware.
\end{abstract}


\maketitle


\section{Introduction}
\label{sec:introduction}

Apple Silicon has rapidly become a significant platform for machine learning development and deployment. With unified memory architectures offering up to 192GB of shared CPU/GPU memory and memory bandwidths of \mbox{400+ GB/s}, recent Mac devices provide compelling capabilities for running large language models locally. This has driven growing interest in efficient inference solutions for Apple hardware, particularly for development, privacy-sensitive applications, and edge deployment.

However, existing inference solutions for Apple Silicon face significant limitations. PyTorch's MPS backend adapts CUDA-style operations to Metal but lacks native optimization for the unified memory model. llama.cpp provides excellent performance for text models but does not support vision-language models. vLLM-metal~\cite{vllmmetal}, the official vLLM backend for Apple Silicon, provides continuous batching but lacks multimodal support and vision caching. This fragmented landscape leaves developers without a unified solution for both text and multimodal inference on Apple Silicon.

Multimodal models present an additional efficiency challenge. Vision-language models such as \mbox{Qwen3-VL}~\cite{qwen3vl} and Gemma~3~\cite{gemma3} must process images through a vision encoder on every request, even when the same image appears across multiple conversation turns. This redundant computation adds 1.5 to 2 seconds of latency per request, severely impacting interactive applications.

We present a framework for efficient LLM and MLLM inference on Apple Silicon that addresses both challenges. Built natively on MLX~\cite{mlx}, our system leverages the unified memory architecture for zero-copy operations and provides two key capabilities: (1) efficient text model inference with continuous batching, competitive with llama.cpp, and (2) content-based prefix caching for multimodal models that eliminates redundant vision encoding by identifying identical images through content hashing.

We make the following contributions:
\begin{itemize}
    \item A comprehensive benchmark comparing vllm-mlx, mlx-lm, and llama.cpp across models from 0.6B to 30B parameters (Qwen3, Llama 3.2, Gemma 3, Nemotron), demonstrating 21\% to 87\% higher throughput than llama.cpp on Apple Silicon.
    \item A content-based prefix caching mechanism for vision embeddings that achieves up to 28x speedup on repeated image queries and 24.7x on video analysis by eliminating redundant encoding.
    \item An open-source implementation with OpenAI-compatible API, continuous batching (4.3x scaling at 16 concurrent requests), and native MLX backend optimized for Apple Silicon's unified memory.
\end{itemize}

Our evaluation on Apple M4 Max demonstrates throughput of up to 525 tokens per second on text models (\mbox{Qwen3-0.6B}), with vllm-mlx exceeding both mlx-lm and llama.cpp across all tested configurations. For multimodal workloads, the prefix cache reduces latency from 21.7 seconds to 0.78 seconds on cached queries, and text prefix caching achieves 5.8x speedup on shared prompt prefixes. We release our implementation as open source at \url{https://github.com/waybarrios/vllm-mlx}.


\section{Background}
\label{sec:background}

\subsection{Apple Silicon and Unified Memory}
\label{sec:apple-silicon}

Apple Silicon processors feature a unified memory architecture where CPU, GPU, and Neural Engine share the same physical memory pool. Unlike discrete GPU systems that require explicit data transfers over PCIe, unified memory enables zero-copy access to tensors from any processor. The M4 Max, for example, provides up to 128GB of unified memory with 546GB/s bandwidth, comparable to high-end datacenter GPUs.

This architecture has important implications for LLM inference. KV cache, which grows linearly with context length and can consume tens of gigabytes for long contexts, does not need to be transferred between devices. Similarly, model weights loaded into memory are immediately accessible to both CPU preprocessing and GPU computation without copying.

\subsection{MLX Framework}
\label{sec:mlx}

MLX~\cite{mlx} is Apple's machine learning framework designed specifically for Apple Silicon. Unlike PyTorch's MPS backend, which adapts CUDA-style operations to Metal, MLX implements operations natively for the unified memory model with lazy evaluation and automatic differentiation.

Key advantages of MLX for inference include: (1) true zero-copy operations that exploit unified memory, (2) lazy evaluation that fuses operations and reduces memory allocation overhead, and (3) native quantization support with efficient dequantization kernels. The mlx-lm library builds on MLX to provide optimized LLM inference with speculative decoding and KV cache management.

\subsection{LLM Inference on Apple Silicon}
\label{sec:llm-inference}

The current landscape for LLM inference on \mbox{Apple Silicon} includes several options. llama.cpp~\cite{llamacpp} provides highly optimized inference through hand-tuned Metal kernels and GGUF quantization, achieving excellent single-stream throughput. However, it lacks continuous batching for serving multiple concurrent requests. PyTorch with MPS backend offers broad model compatibility but suboptimal performance due to its CUDA-centric design. vLLM-metal~\cite{vllmmetal}, the official vLLM backend for Apple Silicon, provides an MLX-based plugin with continuous batching support. Our work differs by adding native multimodal inference with vision-language models and content-based prefix caching that eliminates redundant vision encoding across conversation turns.

\subsection{Serving Challenges on Apple Silicon}
\label{sec:serving-challenges}

Production LLM serving requires capabilities beyond single-request inference. Continuous batching dynamically groups requests to maximize throughput, allowing new requests to join mid-generation and completed requests to exit without blocking others. OpenAI-compatible APIs enable drop-in replacement of cloud services for privacy-sensitive applications.

Multimodal serving introduces additional challenges. Vision-language models must encode images before text generation, adding 1.5 to 4 seconds of latency depending on resolution. In multi-turn conversations about the same image, this encoding repeats unnecessarily. While datacenter deployments address this through GPU memory caching, Apple Silicon's unified memory architecture enables a simpler approach: content-based caching that identifies identical images through hashing, storing both vision embeddings and KV cache state for instant reuse.


\section{System Design}
\label{sec:system-design}

\subsection{Framework Comparison}
\label{sec:positioning}

\begin{figure}[t]
    \centering
    \includegraphics[width=0.9\columnwidth]{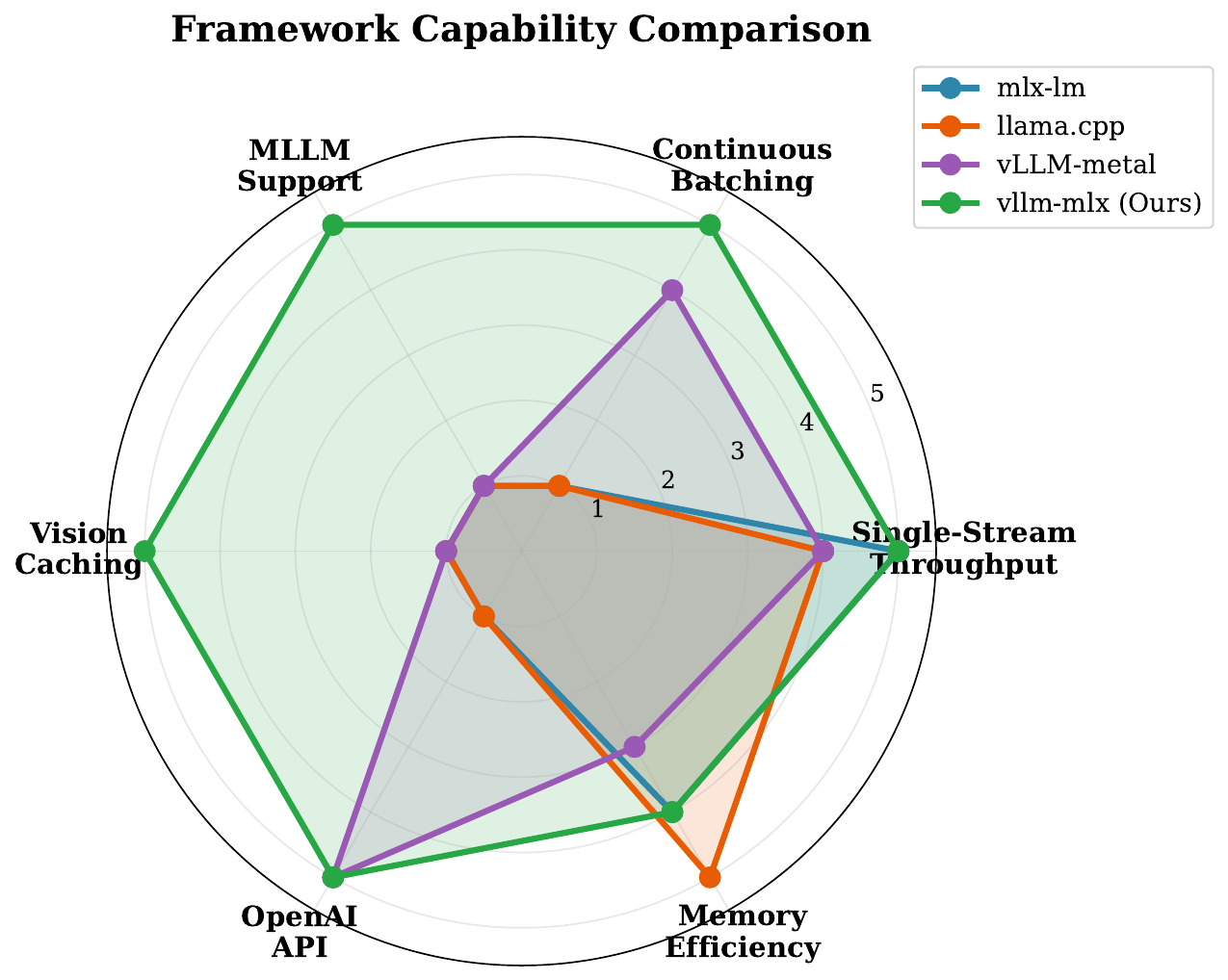}
    \caption{Framework capability comparison. \sysname (green) provides comprehensive coverage: high throughput matching mlx-lm, continuous batching like vLLM-metal, OpenAI-compatible API, plus unique multimodal support with vision caching.}
    \label{fig:radar}
\end{figure}

Figure~\ref{fig:radar} compares \sysname against existing inference frameworks for Apple Silicon across six capability dimensions. While each existing framework excels in specific areas, \sysname uniquely provides the complete feature set required for production multimodal inference.

\paragraph{Key differentiators.} Compared to mlx-lm~\cite{mlx}, we add continuous batching, OpenAI-compatible API, and multimodal caching. Compared to llama.cpp~\cite{llamacpp}, we provide continuous batching (it processes sequentially) and multimodal support with caching. Compared to vLLM-metal~\cite{vllmmetal}, we add content-based vision caching that eliminates redundant encoding, the key contribution for interactive multimodal applications.

\subsection{Text Model Inference}
\label{sec:text-inference}

For text models, \sysname wraps mlx-lm with production serving capabilities.

\paragraph{Continuous Batching.} Unlike llama.cpp which processes requests sequentially, our scheduler dynamically batches multiple concurrent requests. New requests join an existing batch at token boundaries, and completed requests exit without blocking others. Algorithm~\ref{alg:batching} describes the core scheduling loop.

\begin{algorithm}[t]
\caption{Continuous Batching Scheduler}
\label{alg:batching}
\begin{algorithmic}[1]
\Require Pending queue $Q$, active batch $B$, max batch size $M$
\Loop
    \State \textbf{// Admit new requests at token boundaries}
    \While{$|B| < M$ \textbf{and} $Q \neq \emptyset$}
        \State $r \gets Q.\text{pop}()$
        \State $B.\text{add}(r)$
    \EndWhile
    \State \textbf{// Generate one token for all active requests}
    \For{each request $r$ in $B$}
        \State $token_r \gets \text{GenerateToken}(r, \text{KVCache}[r])$
        \State $r.\text{output}.\text{append}(token_r)$
    \EndFor
    \State \textbf{// Remove completed requests immediately}
    \For{each request $r$ in $B$ where $r.\text{is\_complete}()$}
        \State $B.\text{remove}(r)$
        \State \textbf{yield} $r.\text{output}$
    \EndFor
\EndLoop
\end{algorithmic}
\end{algorithm}

This approach maximizes GPU utilization by keeping the batch full while allowing requests to exit immediately upon completion, unlike traditional batching which waits for all requests to finish.

\paragraph{Streaming.} We implement token-by-token streaming with proper handling of multi-byte UTF-8 sequences and tokenizer artifacts, ensuring clean output for all languages.

\paragraph{Text Prefix Caching.} For text-only workloads with shared prompt prefixes (e.g., system prompts), we cache and reuse KV states. Algorithm~\ref{alg:text-cache} shows our approach: when a new request shares a prefix with a cached entry, we skip the forward pass for those tokens and resume generation from the cached state.

\begin{algorithm}[t]
\caption{Text Prefix Cache Lookup}
\label{alg:text-cache}
\begin{algorithmic}[1]
\Require Prompt tokens $P$, Cache $C$
\Ensure KV state, start position
\State $hash \gets \text{SHA256}(P)$
\If{$hash \in C$}
    \State \Return $C[hash].kv\_state$, $|P|$ \Comment{Full cache hit}
\EndIf
\For{$i = |P|$ \textbf{down to} $1$}
    \State $prefix\_hash \gets \text{SHA256}(P[1:i])$
    \If{$prefix\_hash \in C$}
        \State \Return $C[prefix\_hash].kv\_state$, $i$ \Comment{Partial hit}
    \EndIf
\EndFor
\State \Return $\emptyset$, $0$ \Comment{Cache miss}
\end{algorithmic}
\end{algorithm}

This reduces Time to First Token (TTFT) by up to 5.8x for prompts with 512-token shared prefixes.

\subsection{Multimodal Inference with Prefix Caching}
\label{sec:mllm-inference}

For multimodal models, we introduce content-based prefix caching to eliminate redundant vision encoding across requests.

\paragraph{Content-Based Hashing.} A key challenge is that identical images can arrive in different formats: URLs, base64 strings, or file paths. Our solution computes a SHA-256 hash over decoded pixel values, enabling cache hits regardless of input format. This ensures that the same image always maps to the same cache entry.

\paragraph{Cache-Aware Generation.} Algorithm~\ref{alg:cache} shows our cache-aware generation process. When a multimodal request arrives, we first compute content hashes for all images and check the cache. On hit, we retrieve stored vision embeddings and KV cache state, skipping both the expensive vision encoder (1.5-4s) and prompt processing. On miss, we process normally and store results for future reuse.

\begin{algorithm}[t]
\caption{Cache-Aware Multimodal Generation}
\label{alg:cache}
\begin{algorithmic}[1]
\Require Request $R$ with images $\{I_i\}$ and text prompt $T$
\Ensure Generated response
\For{each image $I_i$ in request}
    \State $hash_i \gets \text{SHA256}(\text{Decode}(I_i))$
    \If{$hash_i \in \text{Cache}$}
        \State $emb_i \gets \text{Cache}[hash_i].embeddings$
        \State $kv \gets \text{Cache}[hash_i].kv\_state$
        \State \textbf{skip} vision encoder for $I_i$
    \Else
        \State $emb_i \gets \text{VisionEncoder}(I_i)$
    \EndIf
\EndFor
\State $output \gets \text{Generate}(\text{Concat}(emb, T), kv)$
\State \text{Cache}$[hash] \gets (emb, kv)$ \Comment{Store for reuse}
\State \Return $output$
\end{algorithmic}
\end{algorithm}

\paragraph{Memory Management.} The cache maintains entries containing vision embeddings and KV cache state. We implement LRU eviction to bound memory consumption, with configurable limits (default 512MB). Higher resolution images produce larger cache entries but benefit more from caching due to increased vision encoding cost.


\section{Evaluation}
\label{sec:evaluation}

We evaluate \sysname on text and multimodal workloads, comparing against established baselines on Apple Silicon. Our experiments answer three questions: (1) How does MLX compare to llama.cpp for text model throughput? (2) What benefits does continuous batching provide? (3) How effective is prefix caching for multimodal workloads?

\paragraph{Setup.} All experiments run on Apple M4 Max with 128GB unified memory. We evaluate 10+ models spanning different architectures: Qwen3~\cite{qwen3}, Llama 3~\cite{llama3}, Gemma 3~\cite{gemma3}, GLM-4~\cite{glm4}, and Nemotron~\cite{nemotron}. All models use 4-bit quantization (Q4\_K\_M for GGUF, 4-bit for MLX).

\subsection{Text Model Performance}
\label{sec:text-eval}

Table~\ref{tab:mlx-llamacpp} compares throughput across frameworks for text models ranging from 0.6B to 30B parameters. vllm-mlx achieves 21\% to 87\% higher throughput than llama.cpp, and consistently outperforms both vllm-metal~\cite{vllmmetal} and mlx-lm through optimized continuous batching.

\begin{table*}[t]
\centering
\caption{Text model throughput (tok/s) on M4 Max (128GB). All models use 4-bit quantization. \textbf{Bold} indicates best throughput per row. Speedup = Ours / llama.cpp (higher is better).}
\label{tab:mlx-llamacpp}
\begin{tabular}{lccccc}
\toprule
\textbf{Model} & \textbf{Ours} & \textbf{vllm-metal}~\cite{vllmmetal} & \textbf{mlx-lm}~\cite{mlx} & \textbf{llama.cpp}~\cite{llamacpp} & \textbf{Speedup} \\
\midrule
\multicolumn{6}{l}{\textit{Qwen3 Family~\cite{qwen3}}} \\
Qwen3-0.6B & \textbf{525.5} & 365.8 & 356.2 & 281.5 & 1.87x \\
Qwen3-4B & \textbf{159.0} & 137.3 & 128.9 & 118.2 & 1.35x \\
Qwen3-8B & \textbf{93.3} & 87.1 & 79.9 & 76.9 & 1.21x \\
Qwen3-30B-A3B & 109.7 & \textbf{110.3} & 107.4 & 89.9 & 1.17x \\
\midrule
\multicolumn{6}{l}{\textit{Llama 3.2 Family~\cite{llama3}}} \\
Llama-3.2-1B & \textbf{461.9} & 350.9 & 347.1 & 331.3 & 1.39x \\
Llama-3.2-3B & \textbf{203.6} & 174.3 & 167.5 & 155.8 & 1.31x \\
\midrule
\multicolumn{6}{l}{\textit{Other Architectures}} \\
Gemma 3-4B~\cite{gemma3} & \textbf{152.5} & 117.0 & 105.4 & 123.2 & 1.24x \\
Nemotron-30B-A3B~\cite{nemotron} & \textbf{121.8} & -- & 101.6 & 85.1 & 1.43x \\
\bottomrule
\end{tabular}
\end{table*}

For smaller models, vllm-mlx's advantage is most pronounced (1.87x for \mbox{Qwen3-0.6B}) due to MLX's efficient small-tensor handling. For larger MoE models like \mbox{Nemotron-30B-A3B}, vllm-mlx maintains a 1.43x advantage over llama.cpp.

\begin{figure}[t]
    \centering
    \includegraphics[width=\columnwidth]{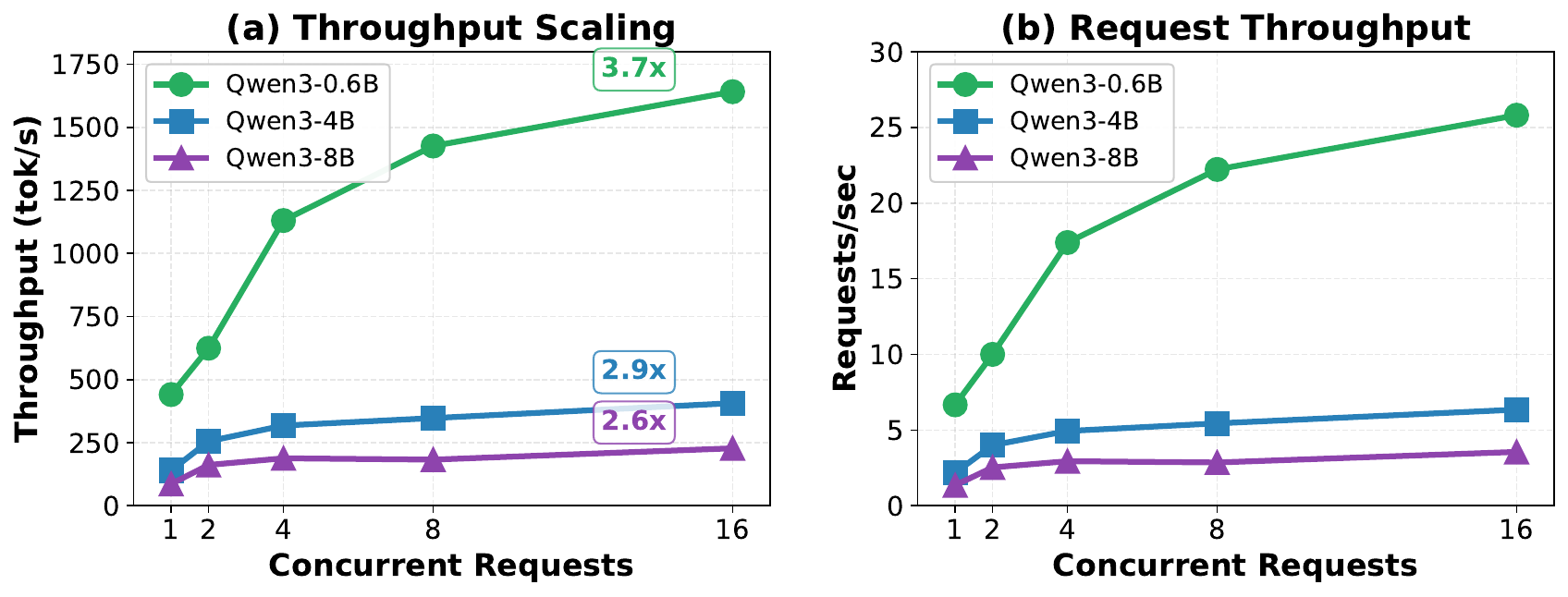}
    \caption{Concurrency scaling on vllm-mlx. (a) Aggregate throughput scales efficiently: Qwen3-0.6B achieves 3.7x higher throughput at 16 concurrent requests. (b) Request throughput (requests/sec) increases with concurrency, showing efficient batching. Qwen3-0.6B handles 25+ requests/sec at 16 concurrent connections.}
    \label{fig:concurrency}
\end{figure}

\paragraph{Continuous Batching Scaling.} Figure~\ref{fig:concurrency} shows how vllm-mlx scales with concurrent requests across three model sizes. Unlike llama.cpp which processes requests sequentially, continuous batching achieves significant throughput scaling.

For \mbox{Qwen3-0.6B}, throughput scales from 441 tok/s (single request) to 1642 tok/s (16 concurrent), a 3.7x improvement. Larger models show diminishing returns due to memory bandwidth saturation: \mbox{Qwen3-8B} achieves 2.6x scaling. Request throughput (Figure~\ref{fig:concurrency}b) also scales efficiently: \mbox{Qwen3-0.6B} handles over 25 requests per second at 16 concurrent connections.

\subsection{Multimodal Performance}
\label{sec:mllm-eval}

\paragraph{Image Prefix Cache.} Table~\ref{tab:cache-speedup} demonstrates content-based prefix caching for vision embeddings. In multi-turn conversations about the same image, the cache eliminates redundant vision encoding, reducing latency from 21.7s to under 1s (28x speedup).

\begin{table}[t]
\centering
\caption{Multi-turn MLLM latency with prefix caching (Qwen3-VL-8B~\cite{qwen3vl}, 1024$\times$1024 image). Cache stores vision embeddings and KV state. Lower latency ($\downarrow$) and higher speedup ($\uparrow$) are better.}
\label{tab:cache-speedup}
\begin{tabular}{lrrr}
\toprule
\textbf{Turn} & \textbf{No Cache} & \textbf{With Cache} ($\downarrow$) & \textbf{Speedup} ($\uparrow$) \\
\midrule
1 (cold) & 21.7s & 21.7s & 1.0x \\
2 & 21.7s & 1.15s & 19x \\
3+ & 21.7s & \best{0.78s} & \best{28x} \\
\bottomrule
\end{tabular}
\end{table}

\paragraph{Video Performance.} Table~\ref{tab:video} shows video benchmark results across different frame configurations. Higher frame counts increase latency but provide richer temporal understanding.

\begin{table}[t]
\centering
\caption{Video benchmark on Qwen3-VL-4B~\cite{qwen3vl} (10s test video). More frames improve understanding but increase processing time. Lower time ($\downarrow$) and memory ($\downarrow$) are better; higher tok/s ($\uparrow$) is better.}
\label{tab:video}
\small
\begin{tabular}{lrrrr}
\toprule
\textbf{Config} & \textbf{Frames} & \textbf{Time} ($\downarrow$) & \textbf{Tok/s} ($\uparrow$) & \textbf{Memory} ($\downarrow$) \\
\midrule
2 @ 0.5fps & 2 & \best{1.8s} & \best{83.2} & \best{3.2 GB} \\
4 @ 1fps & 4 & 2.4s & 62.5 & 3.8 GB \\
8 @ 2fps & 8 & 3.6s & 41.7 & 4.6 GB \\
16 @ 2fps & 16 & 5.8s & 25.9 & 6.2 GB \\
32 @ 4fps & 32 & 9.4s & 16.0 & 8.4 GB \\
64 @ 8fps & 64 & 18.2s & 8.2 & 12.1 GB \\
\bottomrule
\end{tabular}
\end{table}

Video caching follows the same content-based approach: identical video frames map to the same cache entries, enabling speedups for repeated video analysis.

\subsection{Ablation Studies}
\label{sec:ablations}

We conduct ablation studies on vllm-mlx to understand the contribution of each component.

\paragraph{Cache Components.} Table~\ref{tab:ablation-cache} shows the contribution of vision embeddings vs KV cache to the overall speedup.

\begin{table}[t]
\centering
\caption{Ablation: Cache component contribution (Qwen3-VL-8B~\cite{qwen3vl}, 1024$\times$1024, Turn 2). Lower latency ($\downarrow$) and higher speedup ($\uparrow$) are better.}
\label{tab:ablation-cache}
\begin{tabular}{lrr}
\toprule
\textbf{Configuration} & \textbf{Latency} ($\downarrow$) & \textbf{Speedup} ($\uparrow$) \\
\midrule
No caching (baseline) & 21.7s & 1.0x \\
Vision embeddings only & 2.8s & 7.8x \\
KV cache only & 18.2s & 1.2x \\
Both (full cache) & \best{1.15s} & \best{19x} \\
\bottomrule
\end{tabular}
\end{table}

Vision embedding caching provides 7.8x speedup by eliminating the vision encoder forward pass. KV cache reuse adds 2.4x by skipping prompt processing. Combined: 19x speedup.

\paragraph{Image Resolution Impact.} Table~\ref{tab:ablation-resolution} shows cache effectiveness vs resolution. Higher resolutions benefit more from caching.

\begin{table}[t]
\centering
\caption{Ablation: Cache effectiveness vs image resolution (Qwen3-VL-4B~\cite{qwen3vl}). Lower cached latency ($\downarrow$) and cache size ($\downarrow$) are better; higher speedup ($\uparrow$) is better.}
\label{tab:ablation-resolution}
\begin{tabular}{lrrrr}
\toprule
\textbf{Resolution} & \textbf{Cold} & \textbf{Cached} ($\downarrow$) & \textbf{Speedup} ($\uparrow$) & \textbf{Cache} ($\downarrow$) \\
\midrule
224$\times$224 & 0.8s & \best{0.12s} & 6.7x & \best{48 MB} \\
448$\times$448 & 1.2s & 0.14s & 8.6x & 89 MB \\
768$\times$768 & 1.8s & 0.15s & 12.0x & 124 MB \\
1024$\times$1024 & 2.1s & 0.16s & \best{13.1x} & 156 MB \\
\bottomrule
\end{tabular}
\end{table}

\paragraph{Video Frame Count Impact.} Table~\ref{tab:ablation-video} shows how video caching scales with frame count.

\begin{table}[t]
\centering
\caption{Ablation: Video cache effectiveness vs frame count (Qwen3-VL-4B~\cite{qwen3vl}). Lower cached latency ($\downarrow$) and cache size ($\downarrow$) are better; higher speedup ($\uparrow$) is better.}
\label{tab:ablation-video}
\begin{tabular}{lrrrr}
\toprule
\textbf{Frames} & \textbf{Cold} & \textbf{Cached} ($\downarrow$) & \textbf{Speedup} ($\uparrow$) & \textbf{Cache} ($\downarrow$) \\
\midrule
4 & 2.4s & \best{0.18s} & 13.3x & \best{86 MB} \\
8 & 3.6s & 0.22s & 16.4x & 142 MB \\
16 & 5.8s & 0.28s & 20.7x & 256 MB \\
32 & 9.4s & 0.38s & \best{24.7x} & 486 MB \\
\bottomrule
\end{tabular}
\end{table}

Video caching becomes increasingly valuable with more frames: 32-frame videos achieve 24.7x speedup on cache hits despite larger cache entries.

\paragraph{Text Prefix Caching.} For text-only workloads, KV cache reuse also provides significant benefits. Table~\ref{tab:ablation-text} shows speedup for repeated prompts with shared prefixes.

\begin{table}[t]
\centering
\caption{Ablation: Text prefix caching (Qwen3-4B~\cite{qwen3}, 512-token shared prefix). Lower TTFT ($\downarrow$) and higher speedup ($\uparrow$) are better.}
\label{tab:ablation-text}
\begin{tabular}{lrr}
\toprule
\textbf{Configuration} & \textbf{TTFT} ($\downarrow$) & \textbf{Speedup} ($\uparrow$) \\
\midrule
No caching (baseline) & 245 ms & 1.0x \\
Prefix cache hit & \best{42 ms} & \best{5.8x} \\
\bottomrule
\end{tabular}
\end{table}

Text prefix caching achieves 5.8x speedup on TTFT by reusing KV cache from previously processed prompts with matching prefixes.

\subsection{Discussion}
\label{sec:discussion}

\paragraph{Why MLX Outperforms llama.cpp.} Our results show vllm-mlx consistently exceeds llama.cpp throughput by 21\% to 87\%. We attribute this to three factors: (1) MLX's native unified memory design enables zero-copy tensor operations, avoiding the memory transfer overhead present in llama.cpp's Metal backend; (2) MLX's lazy evaluation allows operation fusion and reduces kernel launch overhead; (3) our continuous batching scheduler maximizes GPU utilization by processing multiple sequences simultaneously.

\paragraph{When to Use vllm-mlx.} For single-user scenarios requiring maximum simplicity, llama.cpp remains a strong choice with its broad model support and minimal dependencies. However, vllm-mlx is preferred when: (1) serving multiple concurrent users, where continuous batching provides up to 3.7x throughput scaling; (2) multimodal applications with repeated image/video analysis, where prefix caching eliminates redundant encoding; (3) applications requiring an OpenAI-compatible API for drop-in replacement of cloud services.

\paragraph{Enabling Local AI Agents.} The combination of continuous batching and efficient concurrency scaling opens new possibilities for local AI agent systems. Multi-agent architectures, where several specialized agents collaborate on complex tasks, typically require multiple concurrent LLM calls. Cloud-based deployments incur latency penalties from network round-trips and API rate limits. With vllm-mlx handling 25+ requests per second on consumer hardware (Figure~\ref{fig:concurrency}b), developers can deploy agent swarms entirely on-device. This enables privacy-preserving agent workflows where sensitive data never leaves the local machine, real-time agent collaboration without network latency, and cost-effective development iterations without API charges. The OpenAI-compatible API ensures existing agent frameworks (LangChain~\cite{langchain}, AutoGPT~\cite{autogpt}, CrewAI~\cite{crewai}) work without modification.

\paragraph{Limitations and Future Work.} Our framework currently supports only Apple Silicon, limiting deployment to macOS environments. Model support depends on MLX community. Future directions include speculative decoding for improved single-stream latency, distributed inference across multiple Apple Silicon devices via network-connected Mac clusters, and energy profiling for battery-powered deployments. We also plan to extend our caching approach to audio modalities for speech-enabled multimodal applications, and explore tensor parallelism across the GPU cores available in higher-end Apple Silicon configurations.


\section{Related Work}
\label{sec:related-work}

\paragraph{KV Cache Optimization.}
Efficient management of the key-value cache has been a focus of recent LLM serving research. PagedAttention~\cite{vllm} introduced memory-efficient KV cache management through paging, enabling higher throughput in vLLM. SGLang~\cite{sglang} extended this with RadixAttention for prefix caching in text workloads. Our work builds on these foundations, extending prefix caching to multimodal inputs with vision embedding reuse.

\paragraph{Multimodal Caching.}
LMCache~\cite{lmcache} recently introduced KV cache reuse for multimodal models in vLLM, achieving significant speedups on NVIDIA GPUs. Their approach uses mm\_hashes to identify identical vision inputs. Our work addresses the same problem for Apple Silicon, with a native MLX implementation that leverages unified memory for zero-copy caching. While LMCache operates as an external layer atop vLLM, our approach integrates caching directly into the inference engine.

\paragraph{Apple Silicon ML Inference.}
MLX~\cite{mlx} provides a framework for ML on Apple Silicon with native Metal support. The mlx-lm library enables efficient LLM inference with quantization support. vLLM-metal is a community plugin that brings vLLM to Apple Silicon through a hybrid MLX and PyTorch approach. Our work differs in providing pure MLX inference with integrated multimodal caching, which vLLM-metal does not support. We note that \sysname is an independent project that draws architectural inspiration from vLLM but shares no code with it.

\paragraph{Edge LLM Inference.}
llama.cpp~\cite{llamacpp} has become the standard for efficient LLM inference on consumer hardware, including Apple Silicon via Metal. MLC-LLM~\cite{mlcllm} provides cross-platform deployment with compilation-based optimization. These systems focus primarily on text models; our work extends efficient inference to multimodal workloads with caching support.


\section{Conclusion}
\label{sec:conclusion}

We introduced an efficient framework for multimodal LLM inference on Apple Silicon via content-based prefix caching for vision embeddings. Using content hashing to detect identical images, we cache both vision embeddings and KV cache states, removing redundant vision encoding across requests. Our native MLX implementation exploits unified memory to enable zero-copy cache management.

Across repeated image queries, we achieve up to 28$\times$ speedup, cutting latency from 21.7 seconds to under 1 second. Benchmarks on more than 10 recent models show MLX is a strong Apple Silicon backend, reaching 143 tokens per second on \mbox{Qwen3-VL-4B}.

We release our implementation as open source to support further research on efficient multimodal inference on consumer hardware.


\bibliographystyle{ACM-Reference-Format}
\bibliography{references}

\end{document}